\documentclass[10pt,a4paper,twocolumn]{article}
\ifdefined\XeTeXrevision
\usepackage{fontspec}
\fi
\usepackage[left=17mm,right=17mm,top=20mm,bottom=20mm,columnsep=7mm]{geometry}
\usepackage{microtype}
\usepackage{booktabs}
\usepackage{array}
\usepackage{amssymb,amsmath}
\usepackage{tikz}
\usetikzlibrary{arrows.meta,positioning,calc}
\usepackage{caption}
\usepackage{titlesec}
\titleformat{\section}{\normalfont\large\bfseries}{\thesection}{0.6em}{}
\titleformat{\subsection}{\normalfont\normalsize\bfseries}{\thesubsection}{0.6em}{}
\titlespacing*{\section}{0pt}{10pt}{4pt}
\titlespacing*{\subsection}{0pt}{8pt}{3pt}
\usepackage[colorlinks=true,linkcolor=black,urlcolor=blue!60!black,citecolor=black]{hyperref}
\hypersetup{pdftitle={Grouping the Stochastic Machine},pdfauthor={George Andrikopoulos}}
\usepackage{url}
\usepackage{balance}
\providecommand{\tightlist}{\setlength{\itemsep}{0pt}\setlength{\parskip}{0pt}}
\newcommand{\FigTargets}{%
\begin{figure*}[t]
\centering
\begin{tikzpicture}[font=\footnotesize]
\foreach \x/\lab/\sub in {
  0/{Accurate, precise}/{on target, tight},
  4.4/{Accurate, imprecise}/{centred, scattered},
  8.8/{Inaccurate, precise}/{tight, off-centre},
  13.2/{Inaccurate, imprecise}/{scattered and off}}
{
  \begin{scope}[shift={(\x,0)}]
    \draw[black!20] (0,0) circle (1.45);
    \draw[black!35] (0,0) circle (0.95);
    \draw[black!55] (0,0) circle (0.48);
    \fill[black!70] (0,0) circle (0.05);
    \node[font=\scriptsize\bfseries, align=center, text width=32mm] at (0,-2.15) {\lab};
    \node[font=\scriptsize, text=black!60, align=center, text width=32mm] at (0,-2.62) {\sub};
  \end{scope}
}
\foreach \p in {(0.10,0.12),(-0.08,0.05),(0.05,-0.10),(-0.03,-0.06),(0.12,-0.02)}
  \fill[blue!65!black] \p circle (0.075);
\foreach \p in {(4.4+0.85,0.55),(4.4-0.75,0.40),(4.4+0.15,-0.95),(4.4-0.55,-0.60),(4.4+0.45,0.95)}
  \fill[blue!65!black] \p circle (0.075);
\foreach \p in {(8.8+0.78,0.72),(8.8+0.90,0.60),(8.8+0.70,0.85),(8.8+0.86,0.80),(8.8+0.95,0.68)}
  \fill[blue!65!black] \p circle (0.075);
\foreach \p in {(13.2+1.05,0.75),(13.2-0.35,1.15),(13.2+0.95,-0.85),(13.2-0.95,-0.35),(13.2+0.20,0.15)}
  \fill[blue!65!black] \p circle (0.075);

\node[align=center, font=\footnotesize, text width=175mm] at (6.6,-3.35)
  {The two axes are independent. Capability benchmarks report the first; production reliability depends on the second.
   The library of Paper~1 moves a group onto the target --- it cannot tighten one.};
\end{tikzpicture}
\caption{Accuracy is where the mean shot lands; precision is the size of the group. Frontier models have converged on the first and still differ on the second, which is the axis this paper argues should decide selection.}
\label{fig:targets}
\end{figure*}}

\newcommand{\FigDecision}{%
\begin{figure}[t]
\centering
\begin{tikzpicture}[
  font=\scriptsize,
  box/.style={draw=black!55, rounded corners=2pt, fill=black!4, text width=27mm, align=center, minimum height=6.5mm, inner sep=2pt},
  dec/.style={draw=black!60, fill=blue!6, rounded corners=2pt, text width=30mm, align=center, minimum height=7.5mm, inner sep=2pt},
  act/.style={draw=black!60, fill=black!10, rounded corners=2pt, text width=27mm, align=center, minimum height=7mm, inner sep=2pt},
  ar/.style={->, >=stealth, draw=black!55}
]
\node[box] (run) at (0,4.4) {Run task $N\times$, fresh agents, deterministic scorer};
\node[dec] (p)   at (0,3.1) {$p_i \approx 1$?};
\node[act] (keep) at (3.6,3.1) {Keep};
\node[dec] (cause) at (0,1.7) {Do the failures share \textbf{one cause}?};
\node[act, fill=black!14] (zero) at (-2.6,0.2) {\textbf{Zeroable} \\ write the rule};
\node[act] (rifle) at (2.6,0.2) {\textbf{Not zeroable} \\ change model or temperature};

\draw[ar] (run) -- (p);
\draw[ar] (p) -- node[above, font=\scriptsize] {yes} (keep);
\draw[ar] (p) -- node[right, font=\scriptsize, pos=0.4] {no} (cause);
\draw[ar] (cause.west) -| node[above left, font=\scriptsize, pos=0.25] {yes} (zero.north);
\draw[ar] (cause.east) -| node[above right, font=\scriptsize, pos=0.25] {no} (rifle.north);

\node[font=\scriptsize\itshape, text=black!60, align=center, text width=78mm] at (0,-1.05)
  {The pass rate screens; the causes decide. A low-but-nonzero rate whose failures share one cause is a tight group with a stray shot.};
\end{tikzpicture}
\caption{The corrected procedure. An earlier version of this paper keyed the diagnosis on the pass rate alone; the replication in §4.1 showed that a rate the naive rule called scatter was closed completely by a single rule.}
\label{fig:decision}
\end{figure}}

\newcommand{\FigWorked}{%
\begin{figure}[t]
\centering
\begin{tikzpicture}[font=\scriptsize]
\def\bw{0.62}
\draw[black!35] (0,0) -- (7.4,0);
\foreach \y/\lab in {0/0.0, 1.0/0.5, 2.0/1.0} {
  \draw[black!25, dashed] (0,\y) -- (7.4,\y);
  \node[anchor=east, font=\scriptsize, text=black!60] at (-0.1,\y) {\lab};
}
\foreach \x/\v/\c/\lab in {
  0.9/0.0/black!45/{raw\\(no library)},
  2.7/0.4/black!45/{library\\before rule},
  4.5/2.0/blue!60!black/{library\\with rule},
  6.3/2.0/blue!35!black/{replication\\(later model)}}
{
  \fill[\c] (\x-\bw/2,0) rectangle (\x+\bw/2,\v);
  \node[font=\scriptsize, text=black!70, anchor=south] at (\x,\v+0.06) {};
  \node[align=center, font=\scriptsize, text width=20mm, anchor=north] at (\x,-0.12) {\lab};
}
\node[anchor=south, font=\scriptsize] at (0.9,0.06) {0/5};
\node[anchor=south, font=\scriptsize] at (2.7,0.46) {1/5};
\node[anchor=south, font=\scriptsize] at (4.5,2.06) {5/5};
\node[anchor=south, font=\scriptsize] at (6.3,2.06) {5/5};
\node[anchor=east, font=\scriptsize, text=black!60, rotate=90] at (-0.62,1.0) {pass rate};
\end{tikzpicture}
\caption{One gap, found by measurement rather than authored, closed completely by a single versioned rule --- and replicated on a later default model against the same frozen task and scorer.}
\label{fig:worked}
\end{figure}}

\title{\vspace{-6mm}\textbf{\Large Grouping the Stochastic Machine}\\[1.5mm]
{\large Precision, Not Capability, as the Frontier Metric for AI Systems}\vspace{-2mm}}
\author{George Andrikopoulos\\
{\small Independent researcher, London, United Kingdom}\\
{\small\url{https://github.com/george-andrikopoulos}}}
\date{\small August 2026}

\begin{document}
\twocolumn[
  \begin{@twocolumnfalse}
  \maketitle
  \begin{abstract}
  \noindent
  
Frontier language models are compared, marketed, and benchmarked on capability --- what their best or average output can achieve. I argue this measures the wrong axis. The models have saturated \emph{accuracy}: their mean output lands on the target. What now separates one system from another in practice --- and what determines whether a system is usable for engineering work --- is \emph{precision}: how tightly concentrated their outputs are around that target across repeated, identical requests. Borrowing the marksman's distinction, capability is where the average shot lands; reliability is the size of the group. I make three claims. First, precision, not capability, is the frontier differentiator between systems (Devin, Claude, Grok, and their successors), and the industry's benchmark culture systematically fails to measure it because benchmarks report central tendency, not spread. Second, precision is measurable, cheaply and without circularity, by running a fixed suite of \emph{deterministically scored} tasks many times at fixed temperature and computing the per-task consistency of outcomes --- no model-in-the-loop grader required. Third, the measurement is not merely descriptive but decision-guiding: it separates \emph{consistent} failures (a tight group off-centre, correctable by the operating discipline of Paper 1 --- a sight adjustment) from \emph{scattered} failures (a wide group, correctable only by changing the model or its sampling --- a rifle problem). I define a grouping metric, specify a harness that reuses existing challenge-trial infrastructure, and show how tracking a human-AI pair's grouping over time yields the compounding signal that Paper 1's field study requires. A first real run, since replicated, illustrates both the method and its most important limit: one measured gap was closed completely by a single rule (0/5 → 5/5), while a suite of tasks \emph{authored from the rules themselves} found no value, because a frontier model already embodies explicit good practice --- establishing that a discipline's worth is found by measurement on real work, not constructed from its own rulebook.

  \end{abstract}
  \vspace{4mm}
  \end{@twocolumnfalse}
]

\hypertarget{the-wrong-axis}{%
\section{The wrong axis}
\FigTargets\label{the-wrong-axis}}

Ask which model is ``best'' and the answer arrives in the language of capability: benchmark scores, reasoning depth, context length, the impressiveness of the best output anyone has coaxed out of it. This language describes where the average shot lands. It is silent on the question every practitioner actually lives with: \emph{will it do this again, the same way, the next time I ask?}

The two questions are the marksman's two axes, and they are independent. \textbf{Accuracy} is closeness of the mean to the target --- bias. \textbf{Precision} is the tightness of the grouping --- variance. A rifle can be accurate and imprecise (shots centred on the bullseye but widely scattered) or precise and inaccurate (a tight cluster in the corner). They are not the same property, they are not measured the same way, and --- the point of this paper --- they have diverged in exactly this manner across the frontier. The leading models have converged on accuracy: for a well-specified engineering task, the \emph{best} output of any of them is excellent, and their \emph{average} output is on target. What separates them is the width of the group. One returns a correct, idiomatic solution nine times in ten and something wild on the tenth; another is merely good but almost never wild. For production engineering the second is more valuable, and no capability benchmark will tell you so.

This connects directly to Paper 1's rule to judge the tail, not the average. Here I make the sharper claim: the tail \emph{is} the group, precision \emph{is} the frontier metric, and the failure of the field is that it advertises capability-at-best while its users suffer capability-in-expectation.

\hypertarget{why-skills-presuppose-precision-the-zeroing-argument}{%
\section{Why skills presuppose precision: the zeroing argument}\label{why-skills-presuppose-precision-the-zeroing-argument}}

There is a reason precision, not accuracy, is the property that matters for a system that will be \emph{operated} rather than merely admired. In marksmanship you can zero a tight group: if the shots cluster but land low and left, you adjust the sights and the whole cluster moves onto centre. You cannot zero a scattered group --- there is no adjustment that corrects shots landing in different places, because there is nothing consistent to correct.

Paper 1's operating discipline --- versioned skills, the error loop, ``adjust the sights, not the memory'' --- is sight-zeroing. It moves a grouping onto the target. It therefore \emph{presupposes a precision floor}. A skill can make a precise model accurate; it cannot rescue a scattered one, because a correction written against a scattered error never generalises --- the error does not recur in the same place, so the rule never fires against the same miss twice. This yields a clean division of labour and a clean decision rule:

\begin{itemize}
\tightlist
\item
  The \textbf{model's} job is precision --- a tight group.
\item
  The \textbf{operator's} job, through skills, is accuracy --- zeroing that group onto the target.
\end{itemize}

And therefore: \textbf{before investing in guardrails for a failure, measure whether the failure is consistent.} A consistent failure is a tight group off-centre --- write the skill, zero it. A scattered failure is a wide group --- no skill will help; change the rifle (model, or sampling temperature). The measurement in §4 is what tells the two apart.

\hypertarget{related-work-and-the-precise-claim}{%
\section{Related work and the precise claim}\label{related-work-and-the-precise-claim}}

The ingredients exist. Repeated-sampling evaluation is established: pass@k measures whether \emph{any} of k samples succeeds, and its severe cousin --- the probability that \emph{all} k succeed --- is a consistency measure {[}1{]}. Self-consistency exploits output variance deliberately {[}2{]}. On the artifact side, the instruction libraries whose value this paper sets out to price are now a documented practice: Jiang and Nam characterise the persistent rule files developers write for coding assistants across 401 open-source repositories {[}4{]}, which establishes that such libraries exist at scale but not what they are worth. The reliability and evaluation literature is increasingly aware that single-shot benchmarking overstates deployed performance, and temperature's role in the accuracy-diversity trade-off is well understood. Zhou et al.\ put the sharpest version of this on record: across several model families, scaling and instruction-tuning raised average performance while \emph{lowering} reliability --- larger, shaped-up models declined fewer questions and returned confident, plausible, wrong answers more often, including on items their human supervisors failed to catch {[}5{]}. That is the strongest existing statement of the premise this paper begins from: the mean is not the quantity that matters. What I have not found stated plainly is the \emph{framing} --- that precision is the axis on which frontier systems now differ and should be selected --- coupled to an \emph{operational} consequence: that measured consistency predicts whether the operating discipline of Paper 1 can help at all. My claim is narrow and practical: not that variance is unmeasured in the literature, but that it is the decisive selection criterion in practice, that it is cheaply measurable without a stochastic grader, and that its measurement is a decision procedure for the human-AI operating loop, not merely a leaderboard column.

\hypertarget{measuring-the-group}{%
\section{Measuring the group}\label{measuring-the-group}}

\textbf{The clean move: score deterministically.} Precision cannot be measured with a model-in-the-loop grader, because the ruler would itself scatter and the measurement would inherit the circularity that dooms self-graded evaluation. The escape is to measure grouping only on tasks with a \emph{deterministic, binary} outcome: does it compile, do the tests pass, is the linter clean, does it typecheck. These signals have no model in the scoring loop; they are cheap, repeatable, and beyond dispute.

\textbf{The harness.} Fix a small suite of verifiable tasks T = \{t\_1 \ldots{} t\_m\}. Fix a candidate configuration C --- a model, optionally with the skill library loaded (the \emph{pair}), at a fixed, recorded sampling temperature. For each task, run it N times on fresh, independent agents, recording the binary outcome of each run. This reuses, essentially unchanged, the challenge-trial harness of Paper 1: independent fresh-agent runs of a fixed task, differing only in that the axis of variation is the random seed, not the presence of a rule.

\textbf{The metrics.} For task t\_i let p\_i be the observed pass rate over N runs.
- \textbf{Accuracy} of C over the suite: the mean pass rate, mean(p\_i). How many targets are hit.
- \textbf{Precision / grouping score}: the fraction of tasks that land \emph{decisively} --- p\_i ≥ (1−ε) or p\_i ≤ ε for a small ε (e.g.~tasks passing ≥9/10 or ≤1/10). A high grouping score means outcomes are consistent, whichever way they fall; the mushy middle (p\_i near 0.5) is scatter. Per-task scatter is well summarised by the variance p\_i(1−p\_i), maximal at p\_i = 0.5.
- \textbf{Worst-case / reliability}: min over tasks of p\_i, and the all-pass rate (Π p\_i, if the runs are independent) --- the production-relevant ``does it hold every time'' figure.

\textbf{The decision table} this produces, which is the harness's real output:

\FigDecision

\begin{table*}[t]
\centering
\caption{The corrected decision procedure. The pass rate screens; the homogeneity of the failure causes decides. Reading the failing samples is not optional.}
\label{tab:decision}
\small
\begin{tabular}{@{}p{0.205\textwidth}p{0.205\textwidth}p{0.205\textwidth}p{0.205\textwidth}@{}}
\toprule
\textbf{Task outcome pattern} & \textbf{Failure causes} & \textbf{Diagnosis} & \textbf{Action} \\
\midrule
Passes decisively (p\_i ≈ 1) & --- & On target, tight & Keep \\
\textbf{Fails at any rate below pass} & \textbf{one shared cause} & \textbf{Tight group, off-centre} & \textbf{Zeroable --- write a skill (Paper 1's \emph{relearn} loop)} \\
Fails at any rate below pass & heterogeneous causes & Wide group & Not zeroable --- change model or lower sampling temperature \\
\bottomrule
\end{tabular}
\end{table*}

The middle row is where the operating discipline earns its keep; the bottom row is where no amount of discipline will, and knowing which row a failure is in --- before spending effort --- is the point.

\textbf{The pass rate alone does not decide the row, and an earlier version of this table said it did.} It is tempting to read a mid-range (p\_i) as scatter and a near-zero (p\_i) as a tight off-centre group, and for a first screening pass that heuristic is serviceable. It is not sufficient. Scatter in the sense that matters here is \emph{causal}, not merely statistical: four failures for one identical reason, with one sample slipping through, is a tight group with a stray shot, and a single rule moves the whole cluster. Four failures for four different reasons at the same rate is genuinely wide, and no rule will collect them. \textbf{So (p\_i) is the screening signal; homogeneity of the failure causes is the confirming test.} Read the failing samples before deciding the row. This correction was not anticipated --- it was forced by the replication reported in §4.1, where a failure rate the naive table classed as scatter was closed completely by one rule.

\hypertarget{a-worked-example-measure-find-the-miss-zero-it-re-measure}{%
\subsection{A worked example: measure, find the miss, zero it, re-measure}\label{a-worked-example-measure-find-the-miss-zero-it-re-measure}}

The following is a real run of this harness (six-task Rust suite, N=5, provider-default temperature --- see §7 on why the temperature is recorded but not controlled). The raw model was Opus 4.8.

The suite scored \textbf{accuracy 0.833, grouping 1.000, worst-case 0.000}: five tasks decisive-pass, one decisive-fail (\texttt{p\_i\ =\ 0}), zero scatter. Note the pairing of a \emph{perfect} grouping score with an \emph{imperfect} accuracy --- this is not a contradiction but the ideal operator state. A perfectly tight group with one decisive miss is the best position to be in, precisely because the miss is decisive and therefore zeroable; a scattered model at the same accuracy would be worse, because its misses would not be.

The single miss (a parse-validation task) failed 5/5 for one identifiable cause: every sample declared the required \texttt{OutOfRange} error variant, then parsed the input directly into the target narrow type (\texttt{u16}), so an out-of-range value like ``70000'' overflowed to a \emph{parse} error and returned \texttt{NotANumber} --- leaving \texttt{OutOfRange} unreachable for the value it exists to name. The model reasoned itself into the same collapse every time; one sample's own documentation stated the bug it was committing. That signature --- decisive, reproducible, self-explained --- is the mark of a systematic error a sight-adjustment corrects, not a scattered one.

A single rule was added to the operating library (parse into a type wide enough to \emph{represent} the out-of-range value, then range-check), and the identical frozen task was re-measured:

\FigWorked

\begin{table}[t]
\centering
\caption{The worked example: one gap found by measurement, closed by one rule.}
\label{tab:worked}
\small
\begin{tabular}{@{}ll@{}}
\toprule
\textbf{Configuration} & \textbf{pass rate on the task} \\
\midrule
raw (no library) & 0.000 (0/5) \\
pair, library \textbf{before} the rule & 0.200 (1/5) \\
pair, library \textbf{with} the rule & \textbf{1.000 (5/5)} \\
\bottomrule
\end{tabular}
\end{table}

Same task, same suite hash, same deterministic scorer; the rule was the only variable. \textbf{Δ = +1.000, raw → ruled.} Mechanism was verified rather than assumed: all five re-run samples now parse into the wider type, and the surviving documentation reproduces the rule's reasoning where the raw run's documented the bug. This is the compounding loop of Paper 1 closed and measured end to end --- a gap found by measurement, named as an error class, retired by one versioned rule, and confirmed retired by re-measurement.

\textbf{Replication, and what it corrected.} The result was replicated on a later default model against the same frozen suite and scorer: the unassisted arm returned 0.200 and the library arm 1.000, matching the recorded baseline for that model exactly. The centrepiece therefore is not an artefact of a single N=5 draw. But the replication also falsified a claim this paper made about its own method. At 0.200 the unassisted arm is, on a naive reading of the decision table in §4, \emph{scattered} --- and the table said scattered failures are not zeroable. A single rule nonetheless closed it to 1.000. Inspection of the failing samples resolved the apparent contradiction: every failure had the \emph{same} cause, the narrow parse, with one sample passing by luck. That is a tight group with a stray shot, not a wide one. The table has been corrected accordingly (§4): the pass rate screens, the homogeneity of causes decides. It is worth stating plainly that the instrument caught the oversimplification in its own author's rule, which is the behaviour the method is supposed to have and the reason for reporting replications at all.

A second, quieter lesson came from the same exercise. Two runs on different models each produced a \texttt{1/5} figure --- on one, the unassisted arm; on the other, the library arm --- and a bare \texttt{1/5} quoted in prose was momentarily attached to the wrong pairing. No recorded figure was wrong; the ambiguity lived entirely in citing a number without its run and arm. Cross-run comparisons must therefore be pinned by run identifier \emph{and} arm, never by the bare value. A measurement programme accumulates numbers faster than it accumulates the context that makes them mean something, and the context is the part that decays.

\hypertarget{two-groupings-the-rifle-and-the-zeroed-rifle}{%
\section{Two groupings: the rifle and the zeroed rifle}\label{two-groupings-the-rifle-and-the-zeroed-rifle}}

Run the harness in two configurations and the difference is itself informative. \textbf{Raw model} (no skills loaded) measures the rifle's intrinsic grouping --- the property to select a model on, alongside cost-per-completed-task. \textbf{The pair} (skills loaded) measures the grouping after zeroing --- the property that governs day-to-day work. The gap between them is the measured value of the operating discipline: how much tighter the library makes the group. And tracked \emph{over time}, the pair's grouping on a fixed suite is the compounding signal Paper 1's field study needs --- as the library matures, tasks failing for one consistent cause should migrate to decisive-pass (each a skill that zeroed a consistent miss), and the grouping score should climb. A rising grouping score on a frozen suite is what ``the pair is getting better'' looks like as a number.

\hypertarget{found-not-authored-the-limit-of-a-constructed-suite}{%
\section{Found, not authored: the limit of a constructed suite}\label{found-not-authored-the-limit-of-a-constructed-suite}}

The worked example above closed a gap completely. An attempt to \emph{measure the library's value across the board} produced the more important result, by failing. To price the operating library, one is tempted to construct a targeted suite --- one task per rule, each built to elicit that rule's error class --- and measure the raw→pair delta. Five such tasks were authored, one per major rule (borrow-not-clone, typed-errors, typestate ordering, illegal-states-unrepresentable, and parse-wide --- the last an \emph{authored} instance of the very class the \emph{found} task of §4.1 had exposed). \textbf{All five failed their own admission gate: the raw frontier model already passed every one}, across successive attempts to make them harder, including a probe that stripped a task's requirement to its intent alone (``make misuse impossible for callers'') --- the model still produced the compile-time-safe design unprompted.

The conclusion is a caution for anyone measuring a discipline's value: \textbf{you cannot author your way to a delta.} Constructing tasks from your own rulebook selects for exactly the classes the model already handles, because a capable model has internalised most explicit good practice; the constructed suite therefore measures the model's baseline competence, not the library's marginal contribution, and the delta it reports is structurally near zero regardless of the library's real worth. Value appears only at the \emph{residual} gaps --- the classes even a frontier model gets wrong --- and those are not knowable in advance. The parse-validation miss of §4.1 was \emph{found by measurement} on a representative task, not designed; it was the only task of the whole exercise that discriminated. The practical corollary: to price an operating library, measure it on representative work and mine the decisive failures, rather than assembling a suite from the rules you already wrote. The library's value lives in what you have not yet thought to encode, and only measurement finds that.

This also bounds the enterprise honestly. On a sufficiently capable model, the marginal value of a comprehensive explicit rulebook trends toward zero, because the model already embodies the discipline; the rules that continue to matter are the few that cover residual, model-specific gaps. A skills library is therefore not a monument to be grown, but a small, live, measurement-fed set of patches over the places the current model still scatters or consistently misses --- and it must be re-measured as models change, because a gap one model has is a gap the next may not.

\hypertarget{cost-confounds-and-honest-limits}{%
\section{Cost, confounds, and honest limits}\label{cost-confounds-and-honest-limits}}

The harness is N× the cost of a single evaluation, so the suite stays small and the runs periodic (a quarterly precision review, or a pre-selection gate before adopting a model), never continuous --- subject, like all of Paper 1's machinery, to its own overhead budget.

\textbf{Scorer bias is the dominant threat, and it runs one way.} In building the harness, several scoring defects were found, and every one biased \emph{against} the discipline --- a systematic direction worth naming. Two examples: a forbidden-construct check scanned test code and so penalised the library for writing \emph{more} tests (where \texttt{.expect()} is legitimate); and doc-tests were scored against a template whose crate name could not resolve, failing any documentation the disciplined config produced. A residual case survived: hidden acceptance tests bind exact signatures, so a task's ``pass'' can require a \emph{worse} type than the library correctly chooses (a plain integer where the disciplined output is a non-zero newtype), and the better code fails to typecheck against the fixture. The lesson generalises: \textbf{naive deterministic scorers systematically undercount discipline, because discipline changes the shape of the code in ways a shape-blind scorer punishes.} Correcting two such defects on already-collected data --- with no new runs --- moved a paired grouping score from 0.167 to 0.500; the paired configuration's initial score on that suite (0.167 grouping, apparently far worse than the raw model's) had been mostly instrument, not signal. A representative suite so afflicted prices the model, not the library, and must not be used for the latter.

\textbf{Temperature could not be independently controlled} in the environment used (the tool exposed no temperature parameter); it was recorded honestly as provider-default and unverified rather than invented. This does not void the §4.1 result, but it bounds what the result may claim. Within a single run-pair --- the two arms executed back to back, against the same frozen task and scorer, on the same model --- the sampling configuration is the same unknown constant, so the raw→ruled delta is valid even though the absolute p\_i values are not anchored to a known temperature. \textbf{That guarantee does not extend across run-pairs.} The replication ran on a later default model, whose provider-side default may differ from the earlier one; each pair is therefore internally comparable, and the two pairs are comparable only in the weaker sense that both produced the same outcome. Claims should be stated in exactly that form: deltas within a pair, agreement between pairs, and no cross-run statement about absolute rates.

Three further limits apply. One further caveat belongs with them, since §4's correction introduces it: judging cause-homogeneity requires \emph{reading the failing samples}, which is a human or model judgement and therefore not deterministic in the way the pass/fail scoring is. The screening signal is objective; the confirming test is not. Where the two disagree, the disagreement should be recorded rather than resolved silently. Determinism of scoring constrains the suite to verifiable tasks --- a feature, since those are the tasks the discipline targets, but it means the method says nothing about open-ended generative work, where (per Paper 1's principle on open-ended work, P7) a single target may not exist and scatter may be desirable. Independence of the N runs is assumed; shared caching can correlate them, and the harness must run them genuinely fresh. And N was small in the reported run; the numbers illustrate mechanism, not effect size.

\hypertarget{conclusion}{%
\section{Conclusion}\label{conclusion}}

The industry asks how smart the machines are and has its answer: smart enough. The question that now decides their worth is older and quieter, and every marksman knows it --- not where the best shot lands, but how tight the group. Precision is the frontier metric; it is measurable without a stochastic ruler by counting decisive outcomes over repeated deterministic trials; and the count does not merely rank systems but tells the operator whether a given miss is his to fix with a skill or the rifle's to fix with a change of model. Paper 1 gave the discipline for zeroing the group. This paper gives the instrument for measuring it --- and the two together make a claim the leaderboards cannot: that the unit of performance is not the model's intelligence but the tightness, and the tractability, of its scatter. The first real run bore this out twice over: where a gap existed, one rule closed it from 0/5 to 5/5, measured; and where gaps were \emph{authored} rather than found, the frontier model had already closed them, so the library's value showed as zero --- teaching that a discipline's worth is proven not by cataloguing what a model does well, but by finding, and retiring, the few things it reliably does wrong.

\begin{center}\rule{0.5\linewidth}{0.5pt}\end{center}

\hypertarget{references}{%
\subsection{References}\label{references}}

{[}1{]} Chen, M. et al.~(2021). \emph{Evaluating Large Language Models Trained on Code.} arXiv:2107.03374. (pass@k.)
{[}2{]} Wang, X. et al.~(2022). \emph{Self-Consistency Improves Chain of Thought Reasoning in Language Models.} arXiv:2203.11171.
{[}3{]} Andrikopoulos, G. (2026). \emph{Tuning the Stochastic Machine: A Systems Engineer's Operating Model for Human-AI Engineering.} (Paper 1 --- the operating discipline this measurement serves.)
{[}4{]} Jiang, S. and Nam, D. (2026). \emph{Beyond the Prompt: An Empirical Study of Cursor Rules.} arXiv:2512.18925. To appear, MSR 2026.

{[}5{]} Zhou, L., Schellaert, W., Martínez-Plumed, F., Moros-Daval, Y., Ferri, C., and Hernández-Orallo, J. (2024). \emph{Larger and more instructable language models become less reliable.} Nature 634, 61--68. https://doi.org/10.1038/s41586-024-07930-y

\balance
\end{document}